\documentclass{article}

\PassOptionsToPackage{numbers}{natbib}
\usepackage[final]{neurips_2021}
\usepackage{graphicx} 
\usepackage[whole]{bxcjkjatype}  
\usepackage{subcaption}
\usepackage{listings}
\usepackage{xcolor}
\usepackage{here}
\usepackage{hyperref}
\usepackage{tabularx}
\usepackage{booktabs}
\newcommand{\code}[1]{\colorbox{white}{\texttt{#1}}}

\usepackage[utf8]{inputenc} 
\usepackage[T1]{fontenc}    
\usepackage{amsfonts}       
\usepackage{nicefrac}       
\usepackage{microtype}      
\definecolor{nord0}{HTML}{2E3440}
\definecolor{nord1}{HTML}{3B4252}
\definecolor{nord2}{HTML}{434C5E}
\definecolor{nord3}{HTML}{4C566A}
\definecolor{nord4}{HTML}{D8DEE9}
\definecolor{nord5}{HTML}{E5E9F0}
\definecolor{nord6}{HTML}{ECEFF4}
\definecolor{nord7}{HTML}{8FBCBB}
\definecolor{nord8}{HTML}{88C0D0}
\definecolor{nord9}{HTML}{81A1C1}
\definecolor{nord10}{HTML}{5E81AC}
\definecolor{nord11}{HTML}{BF616A}
\definecolor{nord12}{HTML}{D08770}
\definecolor{nord13}{HTML}{EBCB8B}
\definecolor{nord14}{HTML}{A3BE8C}
\definecolor{nord15}{HTML}{B48EAD}

\lstdefinestyle{nord}{
    backgroundcolor=\color{nord0},
    commentstyle=\color{nord3},
    keywordstyle=\color{nord9},
    numberstyle=\tiny\color{nord3},
    numbers=left,
    numbersep=5pt,
    basicstyle=\ttfamily\small\color{nord4},
    stringstyle=\color{nord14},
    identifierstyle=\color{nord4},
    showstringspaces=false,
    breaklines=true,
    frame=single,
    rulecolor=\color{nord1},
    frameround=ffff
}

\title{Towards Autonomous Hypothesis Verification via Language Models with Minimal Guidance}

\author{Shiro Takagi*, Ryutaro Yamauchi**, Wataru Kumagai** \\
** Independent Researcher \\
*** The University of Tokyo \\
\{takagi4646@gmail.com, ryutaro\_yamauchi@weblab.t.u-tokyo.ac.jp, kumagai@weblab.t.u-tokyo.ac.jp\}
}

\date{September 2023}

\begin{document}

\maketitle

\begin{abstract}
    Research automation efforts usually employ AI as a tool to automate specific tasks within the research process. To create an AI that truly conduct research themselves, it must independently generate hypotheses, design verification plans, and execute verification. Therefore, we investigated if an AI itself could autonomously generate and verify hypothesis for a toy machine learning research problem. We prompted GPT-4 to generate hypotheses and Python code for hypothesis verification  with limited methodological guidance. Our findings suggest that, in some instances, GPT-4 can autonomously generate and validate hypotheses without detailed guidance. While this is a promising result, we also found that none of the verifications were flawless, and there remain significant challenges in achieving autonomous, human-level research using only generic instructions. These findings underscore the need for continued exploration to develop a general and autonomous AI researcher.
\end{abstract}

\section{Introduction}
Throughout history, humanity has advanced by producing knowledge and developing new technologies through research. Since the inception of artificial intelligence (AI) research, one of the primary goals has been to develop AI capable of conducting such research \cite{langley1987scientific,lindsay1993dendral}. In recent years, with the advancement of machine learning, AI has addressed significant scientific problems \cite{wang2023scientific,xu2021artificial,zhang2023artificial}.

However, the realization of an AI that can autonomously conduct research remains an open problem. Many attempts to automate research with AI have primarily employed AI as a tool to solve specific tasks within the research process. Even when AI is tasked with generating or verifying hypotheses, humans often provide the hypothesis candidates in advance or give guidance on verification methods.

For an AI to genuinely excel in research, it should autonomously generate and verify hypotheses without human guidance in methodology or pre-supplied hypothesis candidates. Achieving such an autonomous artificial researcher is challenging due to its inherent technical complexity.

As an initial step towards this objective, we conducted preliminary research \footnote{
GitHub: \url{https://github.com/t46/mock-pipeline}
}. In this study, we investigated if current AI can autonomously generate and verify hypotheses without extensive methodological guidance for a simplified research problem. We posed a toy machine learning research problem to GPT-4 \cite{openai2023gpt}, asking it to produce hypotheses, devise a verification plan, and convert this plan into executable Python code. We intentionally minimized instructions on hypothesis generation, verification, and problem-specific preparations to assess if large language models (LLMs) can autonomously generate and verify hypotheses.

Our findings indicate that GPT-4 can, in a few cases, autonomously generate and validate hypotheses without explicit instructions. The transition from hypothesis generation to the creation of verification plans was generally successful. Considering the complexity of the challenge we addressed, these outcomes seem promising. However, the results were more akin to what might be termed ``toy models'' of research, and none achieved the perfection of human-conducted research. Notably, GPT-4 encountered difficulties when converting the verification plan into Python code.

These findings suggest that there is potential for AI to autonomously conduct research without detailed instructions. However, numerous challenges persist in realizing such an AI. We will delve into these challenges in the subsequent sections of this paper.

\section{Method}

\subsection{Overview}

\begin{figure}[htb]
    \centering
    \includegraphics[width=\linewidth]{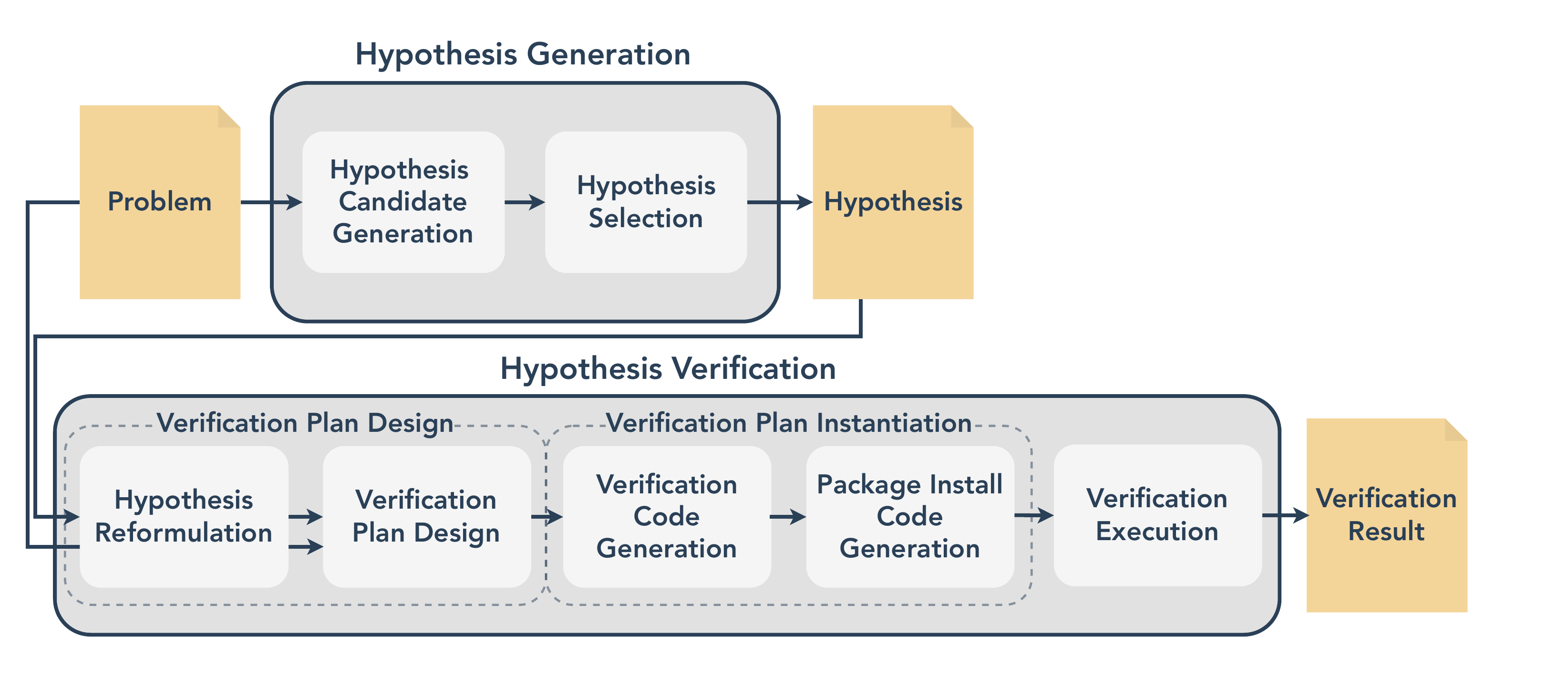}
    \caption{Conceptual diagram of the research pipeline. Upon inputting a research problem into the hypothesis generation module, hypotheses candidates are generated and one hypothesis is selected. This hypothesis is fed into the hypothesis verification module, where it undergoes a reformulation. Based on both the reformulated hypothesis and the problem, a verification plan is devised. This plan guides the generation of Python code for verification and any required package installations. The scripts are then executed to complete the verification.}
    \label{fig:pipeline}
\end{figure}

Our system comprises two major modules: the hypothesis generation module and the hypothesis verification module. For simplicity, we will refer to this system as the \textit{research pipeline}. An overview of the research pipeline is depicted in Figure \ref{fig:pipeline}.

First, we formulated a research problem and input it into the hypothesis generation module as text. Upon receiving the problem, this module generates a hypothesis, also in text form. This hypothesis is then fed into the hypothesis verification module, which produces Python code for verification. Once this code is generated, it is automatically executed by a script we wrote. 

Each module is comprised of several sub-modules. For instance, the hypothesis generation module consists of both the hypothesis candidate generation sub-module and the hypothesis selection sub-module. Each of these sub-modules is powered by GPT-4, which takes the output from the preceding sub-module and a prompt of instruction as its input.

\subsection{Problem}
\label{section:problem}
In our study, we presented GPT-4 with a research problem and asked it to generate hypothesis. Our main goal was to determine if GPT-4 could autonomously handle both hypothesis generation and verification for a simple problem. We tested this using a toy research problem of machine learning, ``LLMs sometimes produce sentences not directly pertinent to the answer.''

For example, when asked ``What is 1 + 1?'', the LLM often replies with ``The answer is 2.'' The ideal response should be the succinct ``2'', rendering the ``The answer is'' portion superfluous. This highlights the issue of the LLM generating sentences that aren't directly pertinent to the answer. This is a challenge particularly for model output evaluation. If the output is ``The answer is 2'' and the correct answer is ``2'', this would be judged as a failure upon direct comparison.

We composed an in-depth description of this problem, supplemented with explanation of why it's problem, and provided it to GPT-4. The specific content given to GPT-4 can be seen in Figure \ref{fig:problem}.

\subsection{Prompts}

\subsubsection{Hypothesis Generation Module}
In the hypothesis generation module, we input the research problem text, as described in Section \ref{section:problem}, into GPT-4 and prompt it to formulate hypotheses. Initially, GPT-4 is instructed to generate multiple hypothesis candidates, shown in Figure \ref{fig:pipeline} as \textit{Hypothesis Candidate Generation}. We then direct it to select the most feasible hypothesis from these candidates, represented in Figure \ref{fig:pipeline} as \textit{Hypothesis Selection}. The selected hypothesis is subsequently passed to the hypothesis verification module.

The prompt for hypothesis candidate generation is depicted in Figure \ref{fig:prompt-hypothesis-candidates-generation}, while that for hypothesis selection can be found in Figure \ref{fig:prompt-hypothesis-selection}. Words enclosed by \textcolor{red}{\{\}} indicate where outputs from earlier sub-modules are inserted. For example, the \textcolor{red}{\{hypotheses\}} in Figure \ref{fig:prompt-hypothesis-selection} includes sentences produced by GPT-4 using the prompt from Figure \ref{fig:prompt-hypothesis-candidates-generation}. For \textcolor{red}{\{problem\}}, we input texts shown in Figure \ref{fig:problem}.

From the prompt, it's evident that we offered only general guidelines. There's no mention of specific problem details, distinct methods for hypothesis generation, or detailed information about hypothesis candidates. This aligns with our prior decision to avoid prescribing a specific method. These guidelines are adaptable and can be applied to hypothesis generation for a range of problems, not just the one in focus. Consequently, we label these instructions as ``general.'' All prompts for the following modules adhere to this core principle.

Indeed, our criterion for hypothesis selection, which is based on ease of verification, has an element of subjectivity. However, in the process of formulating hypotheses, researchers frequently assess their feasibility in relation to available resources. This approach is common across various research domains. Given this backdrop, we believe our instruction maintains a suitable level of generality.

\subsubsection{Hypothesis Verification Module}
\label{section:hypothesis-verification-module}

The hypothesis verification module evaluates the hypothesis. It is divided into three distinct phases. First is the design of the verification plan, represented in Figure \ref{fig:pipeline} as \textit{Verification Plan Design}. Next is the preparation for verification, highlighted in Figure \ref{fig:pipeline} as \textit{Verification Plan Instantiation}. The last phase involves the execution of the verification plan, showcased in Figure \ref{fig:pipeline} as \textit{Verification Execution}.

\subsubsubsection{\textbf{Verification Plan Design}}
\label{section:prompts-verification-plan-design}

When designing the verification plan, GPT-4 uses both the hypothesis and the problem as inputs, generating a textual verification plan. We found that directly creating a verification plan from the raw hypothesis often yielded overly general plans. To counter this, we added a step where GPT-4 ``reformulates'' the hypothesis. This reformulation involves converting the text-based hypothesis into mathematical notation or a similar structured format. Thus, the verification plan design consists of two main steps: hypothesis reformulation (shown in Figure \ref{fig:pipeline} as \textit{Hypothesis Reformulation}) and verification plan construction (depicted in Figure \ref{fig:pipeline} as \textit{Verification Plan Design}).

Figure \ref{fig:prompt-hypothesis-reformulation} presents the prompt used for hypothesis reformulation, while Figure \ref{fig:prompt-verification-design} showcases the prompt for verification plan generation. In the hypothesis reformulation prompt, instead of specifying the exact type of formulation, we guide GPT-4 to conceive the formulation GPT-4 deems appropriate.

In the design of the verification plan, just as with hypothesis selection, we direct the GPT-4 to craft a plan using the easiest method available. Additionally, as previously highlighted, we expect that verification can be achieved through Python code execution. As such, we instructed GPT-4 to ensure the output is executable by language models and computers. 

\subsubsubsection{\textbf{Verification Plan Instantiation}}

To execute verification on a computer, the verification plan must be an computer-executable format. Thus, we instruct GPT-4 to transform the verification plan into a Python script, as depicted in Figure \ref{fig:pipeline} under \textit{Verification Code Generation}. This code is expected to encompass all necessary verification tasks, such as data collection, metric definition, and data analysis. As this phase turns the abstract plan into executable code, we label it \textit{verification plan instantiation}. We've set up a process to extract Python code segments from GPT-4's output and save them as a Python script.

Following the Python code generation (between \textit{Verification Code Generation} and \textit{Package Install Code Generation}, as shown in Figure \ref{fig:pipeline}), we added an additional step to provide GPT-4 with further guidelines. Specifically, we emphasized the avoidance of specifying API key within the code. Instead, we defined API key in advance, which we expected will be used. We also directed GPT-4 to ensure the content was complete, avoiding endings with mere comments or placeholders. Our guidelines were designed to be universally relevant for any API usage, regardless of the specific API.

Setting the API key is problem-dependent. While there are no prompts that include instructions to use specific APIs, we defined the OpenAI API key locally to enable GPT-4 to use another LLM. This decision was made because it wasn't realistic to have GPT-4 set up the API key from scratch, and we deemed it risky. This doesn't truly make the process fully autonomous, so need to be addressed.

While an executable Python script may be generated, it remains non-executable if the requisite packages are not installed. To address this, we prompted the GPT-4 to generate a script that installs the necessary packages to run the code. This is depicted in Figure \ref{fig:pipeline} under \textit{Package Install Code Generation}. Analogous to the verification code, we implemented a procedure to extract Python code segments from the GPT-4's output and save them as a Python script.

In essence, the verification instantiation process encompasses: 1. Generating the verification code, 2. Modifying the code to follow instructions, and 3. Generating the package installation code. The specific prompts used for each of these steps can be found in Appendix \ref{appendix:prompt-hypothesis-verification}. Figure \ref{fig:prompt-verification-code-generation} illustrates the prompt for verification code generation, Figure \ref{fig:prompt-instruction-following} displays the prompt for code modification to adhere to instructions, and Figure \ref{fig:prompt-package-install} showcases the prompt for package installation.

\subsubsubsection{\textbf{Verification Execution}}

We execute the verification by running the verification code saved in a designed directory. We have implemented a procedure to run this code.

If an error occurs during the verification code's execution, we provide GPT-4 with the error message and instruct it to modify the code. In this study, we restrict this correction to errors from the initial run. Even if the revised code still produces an error, we terminate the process without further iterations. The prompt used to modify the code based on the error is depicted in Figure \ref{fig:verification-code-update}.

\subsection{Experiment}
We ran the research pipeline 50 times using the same problem and prompts, subsequently evaluating the outcomes. For each sub-module, we utilized \code{GPT-4} \cite{openai2023gpt} available through the OpenAI API \cite{openaiapi} as of September 2, 2023. Notably, even when the temperature parameter is set to 0, GPT-4 can yield varying outputs. We, therefore, fixed the temperature parameter at zero. This inherent variability stems from the internal mechanics of the API and isn't controlled by an external random seed. 

Given that this is a preliminary study, we employed a rough evaluation strategy. An author reviewed the results, subjectively gauging the suitability of the generated hypothesis and its verification. A more structured and thorough evaluation is planned for future research. We will present a sample output in Section \ref{section:result} and briefly discuss how the author evaluated the sample with supplemental explanations in Appendix \ref{section:sample-analysis} \footnote{
All generated results are on GitHub: \url{https://github.com/t46/mock-pipeline}.
}. We also evaluated the generated code for its executability. If any revisions were made for verification code due to errors, the revised versions were subject to evaluation.

Our evaluation items include: 1. Appropriateness of the hypothesis, 2. Suitability of the verification plan, 3. Appropriateness of the verification code, and 4. Executability of the verification code. Here, ``appropriateness'' refers to the validity of items in relation to their intended purpose. For example, a verification plan's validity in relation to the hypothesis it aims to verify. Even if a hypothesis or verification effort seems basic or lacks originality, if it's free from glaring errors and is a valid response to the question or legitimate verification attempt, it's deemed appropriate.

\section{Results and Discussion}
\label{section:result}

\subsection{Overview}
\label{section:results-overview}

Out of the 50 trials, all were deemed to have generated suitable hypotheses for the presented problems. Among these, 46 were considered feasible given the resources available to the GPT-4. Consequently, appropriate hypothesis generation was achieved in 46 out of the 50 trials. Notably, all 46 of these cases proposed hypotheses along the lines of ``the problem could be resolved by modifying the prompt.'' This hypothesis was consistently selected whenever it appeared among the candidate hypotheses. The remaining four trials, deemed less feasible, all proposed hypotheses suggesting ``the problem could be addressed by training the model.''

From the 50 trials, the majority produced a verification plan that was judged as reasonably appropriate, devoid of any glaring errors. Of these, 24 trials generated Python code deemed somewhat suitable for hypothesis validation. Out of these, 17 trials produced a fully executable Python script. Moreover, 13 of these trials successfully generated code for necessary package installations. Thus, out of the 50 trials, 13 trials successfully generated appropriate verification code.

The 13 trials that generated suitable validation code are encompassed within the 46 trials where the hypothesis was deemed feasible. Therefore, about 25\% of all trials successfully and autonomously executed the entire process, from hypothesis generation to verification code generation. Note that the evaluation is subjective, so please consider the results and specific figures just as a rough guide.

In summary, our results show that GPT-4 can autonomously generate and verify hypotheses using general instructions in a few cases. Given the task's complexity, this is a promising result. However, success was achieved in only 25\% of the trials. While hypothesis and verification plan generation was mostly successful, generating verification code posed challenges. The successful outcomes often resembled prototypes with a shallow understanding of verification rather than human-conducted research. We'll explore these findings further with examples in the next section and Appendix \ref{section:sample-analysis}.

\subsection{Generated Results}
\label{section:results-and-discussion-generate-results}
We will present a specific output example to elucidate our findings and evaluation. We'll explain \code{hypothesis.txt} and \code{verification\_code\_updated.py} in the \code{2023-09-0\_15-57-51} directory in the \code{outputs} folder on GitHub. All examples are available for review on the repository.

\subsubsection{Generated Hypothesis}
The hypothesis produced by the hypothesis generation module is depicted in Figure \ref{fig:result-hypothesis}. As discussed in Section \ref{section:results-overview}, in the majority of cases, the proposed hypothesis was to address the problem by modifying the prompt, as illustrated in this figure. In this particular instance, the recommendation is to append the phrase ``Provide a one-word answer.'' In other instances, the LLM suggested additions like ``Provide the numerical answer'' or proposed rephrasing the prompts to be more specific.

All of these hypotheses were deemed appropriate as they presented plausible solutions to the problem. However, it's evident that they were influenced by the examples provided in the research problem statement. As depicted in Figure \ref{fig:problem}, the text explains the problem using the example ``What is 1 + 1?''. As a result, instead of addressing the broader issue of the LLM generating superfluous outputs, there were instances where it proposed hypotheses like ``Provide the numerical answer'', which seems tailored specifically to this example. This susceptibility to being overly influenced by specific examples is a challenge that warrants attention in future endeavors.

\subsubsection{Generated Verification Code}
We will show an example of the generated verification code in Listing \ref{lst:result_verification_code}. For readability, appropriate line breaks and backquotes have been added.
\textbf{\subsubsubsection{Data Collection}}

First, Listing \ref{lst:result_verification_code} shows that GPT-4 autonomously defines several sample question data. In conventional research, verification based solely on a few sample data would not be deemed persuasive. However, given that we neither provided specific instructions regarding the data nor prepared the dataset for GPT-4 in advance, generating a few samples for verification appears to be a successful results.

Subsequently, they import LLMs and have them generate responses to the questions. Using these unaltered questions as a control group, GPT-4 prepared an experimental group by appending their proposed prompting ``Provide a one-word answer:'' to them. 

Here, there are two things worth noting. The first is that GPT-4 autonomously utilizes the \code{openai} library. Since LLMs can do lots of tasks, ability to operate them indicates the potential for automating processes in numerous research domains. The currently employed API encompasses content only up to 2021; hence, they are using the \code{text-davinci-002} engine, which may not be the most efficient. Nonetheless, this limitation will certainly be addressed as GPT-4 is refined with more recent data.

\textbf{\subsubsubsection{Control Experiment and Verification}}

\label{section:control-experiment-and-verification}

The second is that they utilize the concept of control experiment naturally. Regardless of content quality, code for such a control experiment was produced in the majority of the 50 cases. This is an appropriate approach for verifying their hypothesis, indicating that the GPT-4 has indeed attempted proper verification. Control experiments are a widely accepted verification method in various research domains. Thus, the ability to autonomously employ control experiments without human intervention is promising for the development of AI capable of verification.

GPT-4 compares experimental groups against control groups using verification criteria it designed. In the given example, it assesses word count in each group's outputs to gauge the conciseness of the proposed method. If the ratio of concise responses surpasses a set threshold (here, 0.5), the hypothesis is deemed supported.

Throughout our study, GPT-4 conceived several verification criteria, such as assessing if the output strictly aligns with the answer, its conciseness, whether it's a single word or a numerical value, and its specificity. To verify their hypothesis, they have to ascertain that the output exactly aligns with the answer. Strictly speaking, the concise response, for example, doesn't inherently imply the elimination of irrelevant content. However, we judged the first four were appropriate evaluation metrics for this study. This is because they validate the consequence derived from the hypothesis, a practice that is also common in human-led research.

This verification criteria problem above likely emerges from the process of formulating hypothesis. As depicted in Figure \ref{fig:result-hypothesis}, the term ``concise'' in the prompt can be interpreted diversely. Consequently, when reformulating the hypothesis based on this term, it might equate ``concise'' with ``short in length,'' for example. This discrepancy can potentially be addressed by prompting the GPT-4 to produce more detailed hypotheses or by referencing the problem statement during hypothesis formulation. 

The example of Figure \ref{fig:result-hypothesis} naively contrasts the experiment group with the control group. Yet, many generated results in our trials attempted even statistical hypothesis testing. The capability to autonomously conduct statistical hypothesis testing alongside control experiments is a promising step towards autonomous verification. However, the sample size is often insufficient for most hypothesis tests. Moreover, various prerequisites for each test, such as ensuring Gaussian distribution adherence, aren't checked. Thus, the GPT-4's current use of statistical hypothesis testing may be inappropriate. Teaching the LLMs to grasp human verification methods from foundational principles remains a challenge for future AI development.

Some generated codes were found unsuitable for verification. These can be grouped into two main categories: codes with unsuitable verification criteria and those that simply produce placeholders or comments, such as \code{\# Add your questions here}. Even with explicit instructions to avoid such outputs, this persist. This issue likely stems from the GPT series not being designed to autonomously conduct tasks. If so, addressing these challenges may require rethinking the pre-training phase.

\subsubsubsection{\textbf{Execution}}

One of the most common errors was related to the OpenAI API. The majority of the generated codes utilized \code{openai.Completion.create} or \code{pipeline} and \code{GPT2LMHeadModel} from the \code{transformers} library \cite{wolf2020transformers} for sentence generation. There were 20 instances of errors associated with the OpenAI API. However, after revising the verification code to address these issues, most were resolved. Given the surge in the OpenAI API user base post-2021, it's anticipated that this challenge will be mitigated in the near future.

All cases related to OpenAI adhered to the directive to omit the API key, ensuring no errors in this regard. However, when the verification code was regenerated without re-emphasizing this instruction, errors surfaced in the revised code. In four cases, while the verification code was correct, it faltered solely due to the addition of the API key. This is a problem that can be resolved soon.

\section{Limitation \& Future Work}
\subsubsubsection{\textbf{Problem Settings}}

The first one relates to our problem setting. We gave GPT-4 a simple toy research problem, deliberately avoiding complexities that might stump the model, such as rigorous math or deep logical reasoning. Our chosen problem also didn't require proposing or training models, web resource sourcing, or intricate validation. It likely only needed basic controlled experiments for verification. However, in real research, these complex operations are essential. Therefore, a future challenge is to develop methods that can autonomously handle these complex tasks using general-purpose techniques.

As mentioned, GPT-4 generated hypotheses as prompt suggestions in 46 out of 50 cases. This was somewhat anticipated since we guided the model in that direction. This is because the hypothesis of prompt suggestion doesn't require complex tasks so GPT-4 might autonomously execute it. The strategy of splitting hypothesis generation into generation and selection of candidates furthered this aim. Hence, while we used generic text that could apply to any problem in theory, it's speculated that these might not easily generalize to other scenarios. A future challenge is to assess their adaptability to various contexts and, if not adaptable, to identify necessary adjustments.

\subsubsubsection{\textbf{Generated Outcomes}}

Secondly, there are concerns regarding the results generated by the GPT-4. Out of 50 cases, it failed to produce appropriate verification code in 37 instances. This suggests that GPT-4 might not have a comprehensive understanding or mastery of the concepts it employs in generating results. 

Moreover, the examples produced by the GPT-4 are notably rudimentary, best characterized as toy models for hypothesis verification. They fall short of what can authentically be termed as research. For instance, as highlighted earlier, they often sidestepped the creation of comprehensive datasets in favor of generating a handful of mock data samples. Bridging this gap to achieve the level of meticulous verification conducted by human researchers remains a challenge for the future.

Another crucial consideration is the originality of the hypotheses. Tackling a known problem with a well-tested hypothesis doesn't qualify as authentic research. In our study, we assessed hypotheses only on their relevance to the problem. In the future, it's essential to expand this evaluation, ensuring hypotheses are not just relevant but also showcase novelty.

Thirdly, from a technical standpoint, there were instances where the LLM did not fully follow instructions. Specifically, when generating code, we instructed it not to leave just placeholders or comments, but there were several instances where it did not comply. Constructing a method to always make it follow instructions is crucial for building a more robust system.

\subsubsubsection{\textbf{Autonomy}}

Fourtly, the issue of autonomy warrants discussion. Although our system has demonstrated a significant degree of autonomy in executing various tasks, there remain several challenges.

Firstly, that we supply the problem to GPT-4 is an issue. Currently, we specify the problem and explain why it's a problem. Yet, finding problems is fundamental to research. If the problem is pre-set, the autonomy of the research is questionable. Future improvements should focus on allowing the LLM to independently craft and pursue a research problem.

Secondly, the extraction of pertinent code sections and the subsequent code execution are currently automated based on scripts pre-authored by humans. The ideal scenario would see the LLM independently determining and executing these segments.

Thirdly, as explained in Section \ref{section:hypothesis-verification-module} the OpenAI API key is locally defined by humans. Both this and the previous challenges might be resolved if the LLM were granted more extensive computer operational capabilities. Intensifying initiatives, such as the recently introduced Open Interpreter \cite{openinterpreter}, could be instrumental in addressing these hurdles.

Lastly, the existing system does not possess the capability to autonomously generate new hypotheses based on verification results, thus preventing a closed-loop research practice. The integration of this feature is relatively straightforward, and its future incorporation is a desirable objective.

\subsubsubsection{\textbf{Methodological Soundness}}

Fifthly, our current evaluation method has lots of limitations. As previously mentioned, it's subjective, conducted solely by an author, and based on a small sample size of 50. This raises concerns about potential bias, mistakes, and ambiguity. While we used this method to present initial findings at a workshop, a more objective and extensive evaluation should be pursued in the future.

Lastly, our verification depends on OpenAI's GPT-4 API, whose internals are not transparent and future availability is uncertain. As noted earlier, outputs can vary even with zero temperature. These are constraints on the reproducibility of our results. While we chose GPT-4 for its state-of-the-art performance, future research should consider a more transparent model to ensure reproducibility.

\section{Conclusion}
We presented GPT-4 with a simple toy machine learning research problem and investigated whether GPT-4 could autonomously execute both hypothesis generation and verification for this problem with limited methodological guidance. As a result, we found that there are cases where GPT-4 can autonomously execute the entire process from hypothesis generation to hypothesis verification. Given the difficulty of the problem, this is a promising result. However, none of these were perfect verifications, and we found that there are still many challenges to autonomously generate research at the level humans conduct using only generic instruction. We believe our results provide the first step towards the goal of realizing a general and autonomous artificial researcher by clarifying the challenges ahead.

\bibliographystyle{unsrt}
\bibliography{main}

\appendix

\section{Problem}
\begin{figure}[htb]
    \centering
    \includegraphics[width=\linewidth]{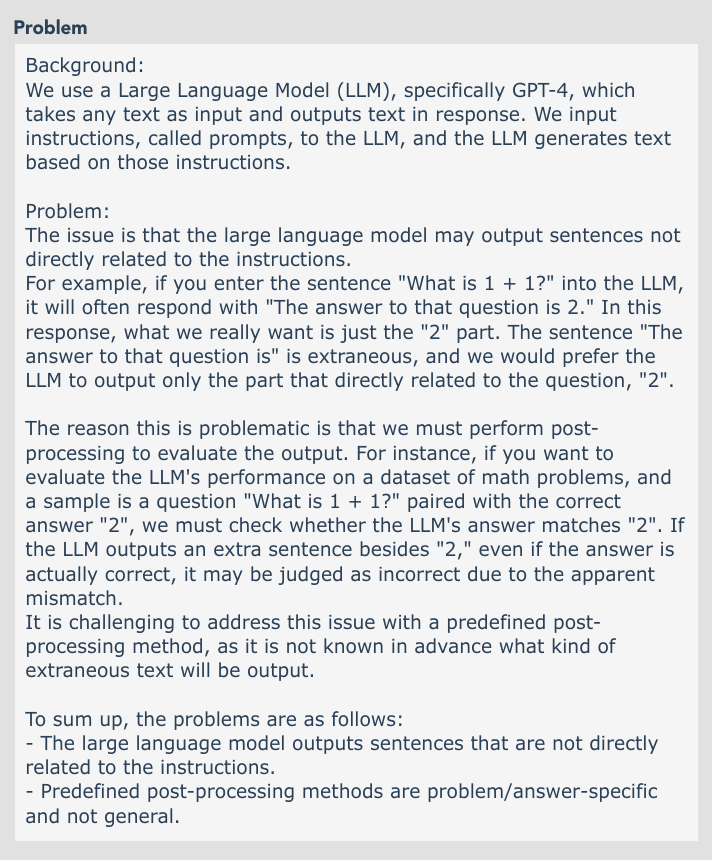}
    \caption{Caption}
    \label{fig:problem}
\end{figure}
\newpage

\section{Prompts}

\subsection{Hypothesis Generation}
\subsubsection{Hypothesis Candidates Generation}
\begin{figure}[htb]
    \centering
    \includegraphics[width=\textwidth]{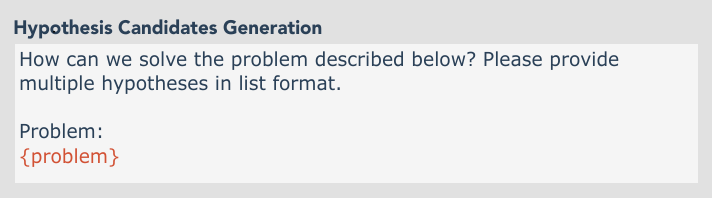}
    \caption{Hypothesis candidates generation}
    \label{fig:prompt-hypothesis-candidates-generation}
\end{figure}
\newpage

\subsubsection{Hypothesis Selection}
\begin{figure}[htb]
    \centering
    \includegraphics[width=\textwidth]{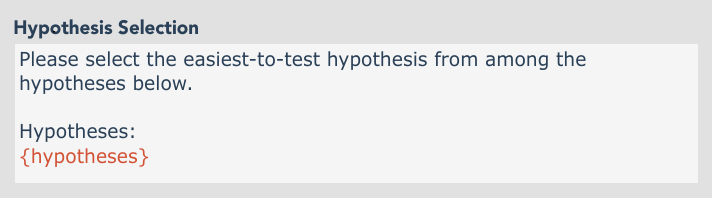}
    \caption{Hypothesis selection}
    \label{fig:prompt-hypothesis-selection}
\end{figure}
\newpage

\subsection{Hypothesis Verification}
\label{appendix:prompt-hypothesis-verification}
\subsubsection{Hypothesis Reformulation}
\begin{figure}[htb]
    \centering
    \includegraphics[width=\textwidth]{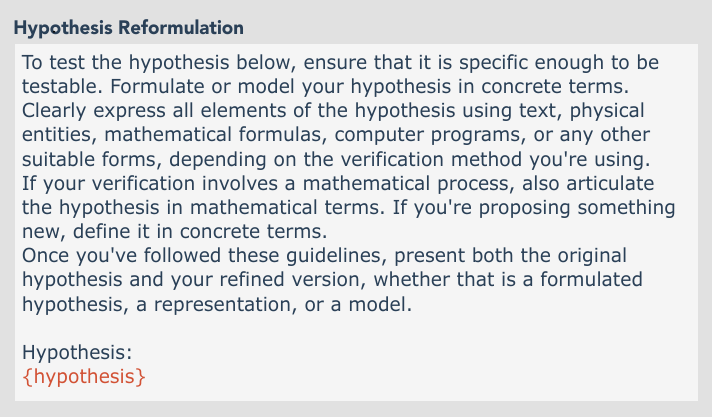}
    \caption{Hypothesis reformulation}
    \label{fig:prompt-hypothesis-reformulation}
\end{figure}
\newpage

\subsubsection{Verification Plan Design}
\begin{figure}[htb]
    \centering
    \includegraphics[width=\textwidth]{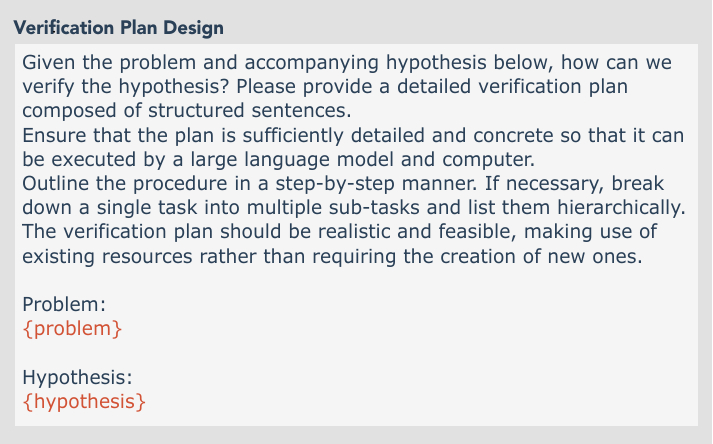}
    \caption{Verification design}
    \label{fig:prompt-verification-design}
\end{figure}
\newpage

\subsubsection{Verification Code Generation}
\begin{figure}[htb]
    \centering
    \includegraphics[width=\textwidth]{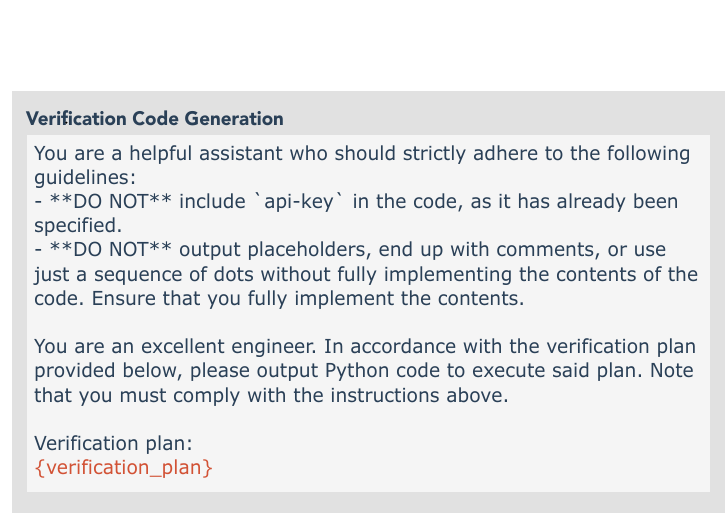}
    \caption{Verification code generation}
    \label{fig:prompt-verification-code-generation}
\end{figure}

\subsubsection{Instruction Following}
\begin{figure}[htb]
    \centering
    \includegraphics[width=\textwidth]{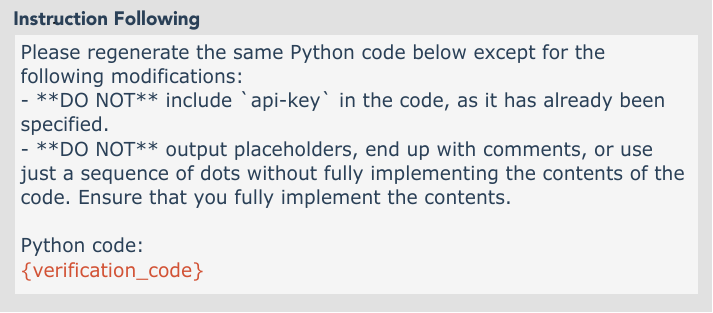}
    \caption{Instruction Following}
    \label{fig:prompt-instruction-following}
\end{figure}
\newpage

\subsubsection{Package Install Code Generation}
\begin{figure}[htb]
    \centering
    \includegraphics[width=\textwidth]{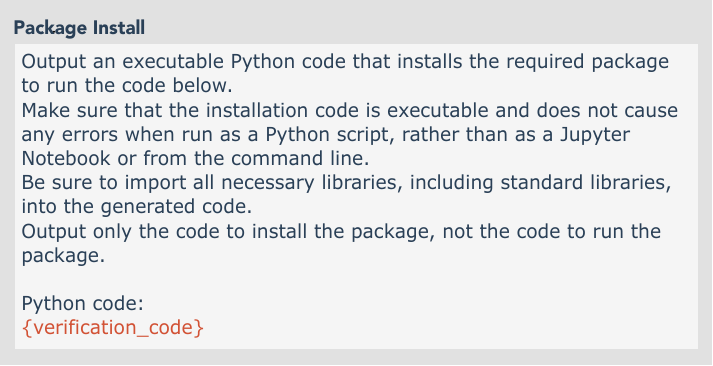}
    \caption{Package install code generation}
    \label{fig:prompt-package-install}
\end{figure}

\subsubsection{Verification Code Update}
\begin{figure}[htb]
    \centering
    \includegraphics[width=\textwidth]{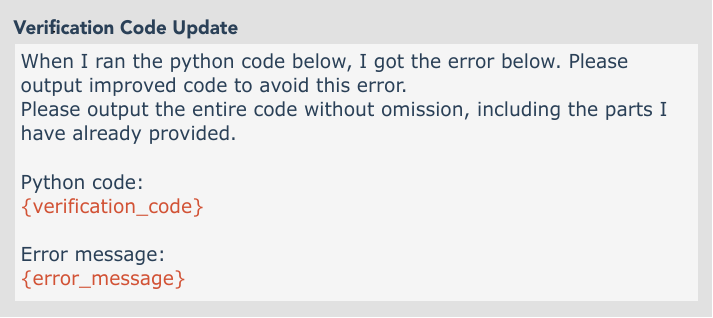}
    \caption{Verification code update}
    \label{fig:verification-code-update}
\end{figure}

\section{Generated Results}

\subsection{Hypothesis Generation}
\subsubsection{Hypothesis Candidates Generation}
\begin{figure}[htb]
    \centering
    \includegraphics[width=\textwidth]{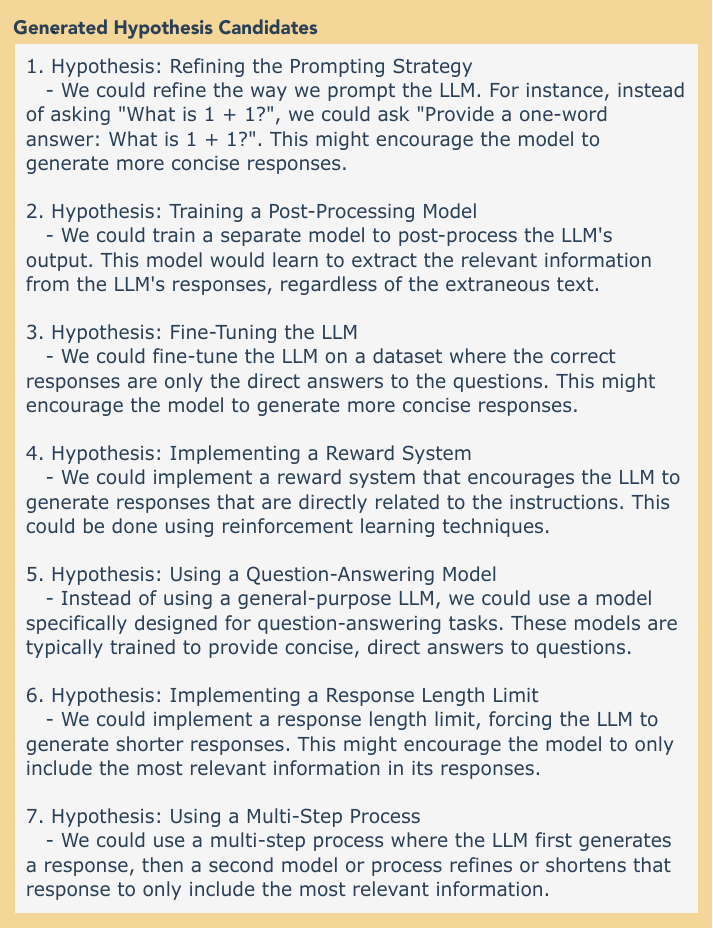}
    \caption{Hypothesis candidates generation}
    \label{fig:result-hypothesis-candidates-generation}
\end{figure}
\newpage

\subsubsection{Hypothesis Selection}
\begin{figure}[htb]
    \centering
    \includegraphics[width=\textwidth]{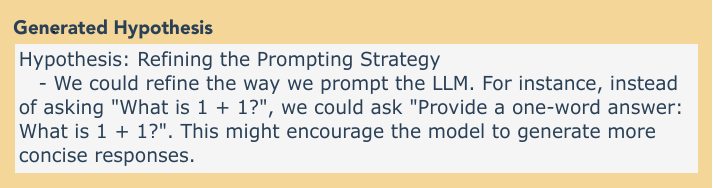}
    \caption{Hypothesis selection}
    \label{fig:result-hypothesis}
\end{figure}
\newpage

\subsection{Hypothesis Verification}
\subsubsection{Hypothesis Reformulation}
\begin{figure}[htb]
    \centering
    \includegraphics[width=\textwidth]{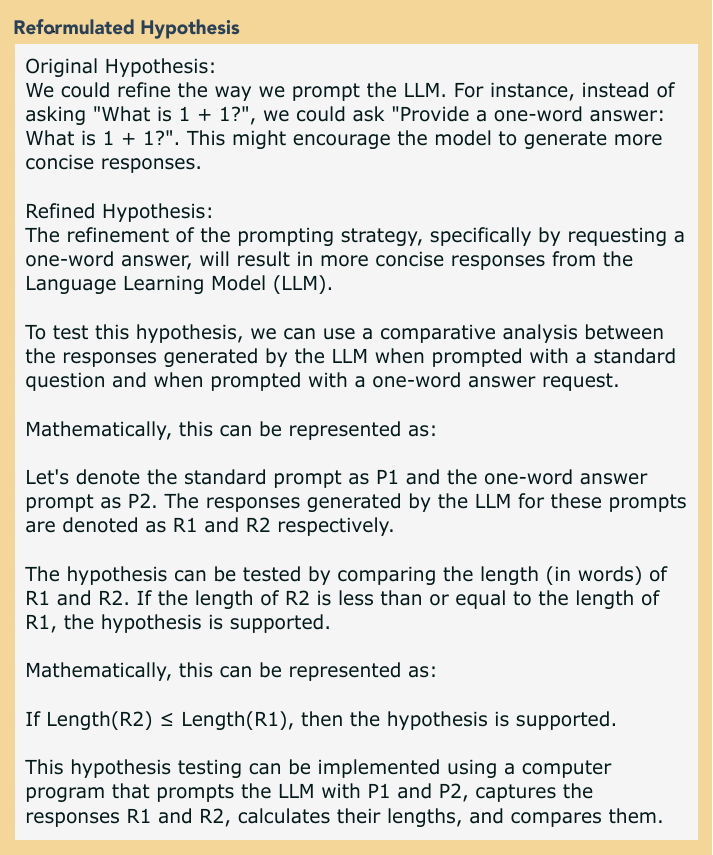}
    \caption{Hypothesis reformulation}
    \label{fig:result-hypothesis-reformulation}
\end{figure}
\newpage

\subsubsection{Verification Plan Design}
\begin{figure}[H]
    \centering
    \includegraphics[width=0.85\textwidth]{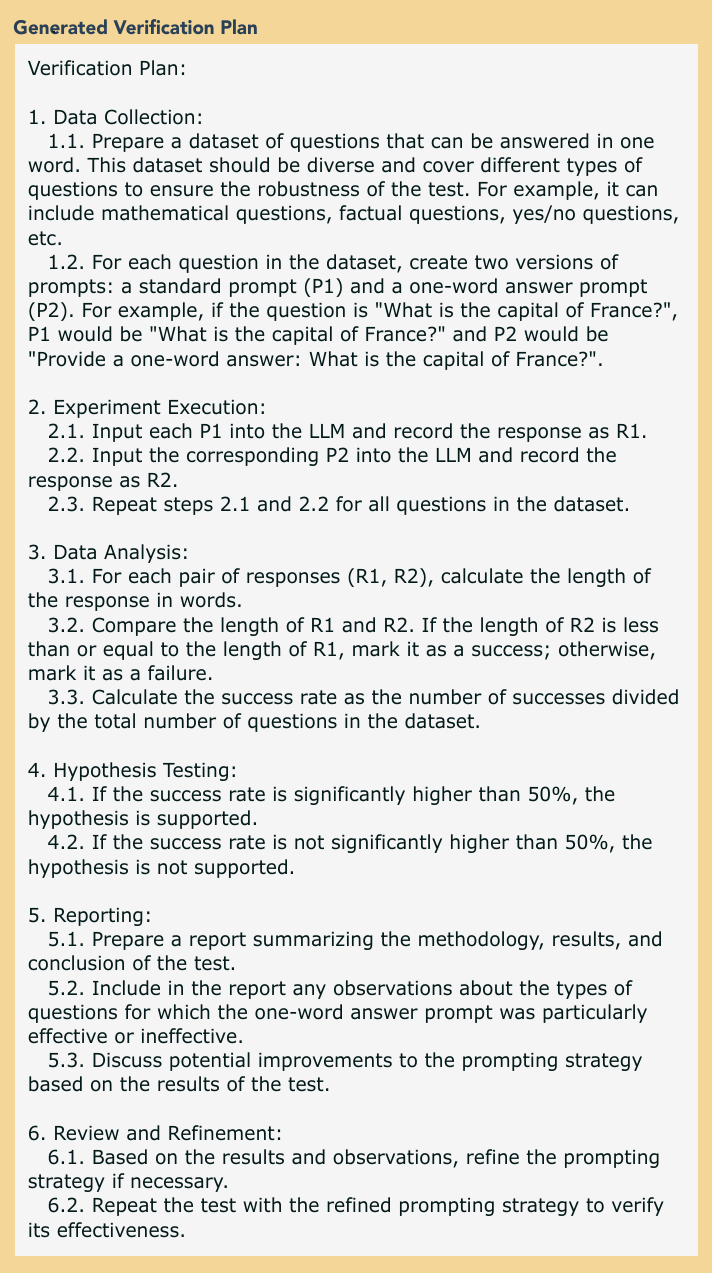}
    \caption{Verification design}
    \label{fig:result-verification-design}
\end{figure}
\newpage

\subsubsection{Verification Code Generation}
\begin{lstlisting}[language=Python,caption={Generated Verification Code},label=lst:result_verification_code]
import openai
import pandas as pd

# 1. Data Collection
questions = ["What is the capital of France?", "What is 2+2?", 
             "Is the sky blue?", "Who wrote 'To Kill a Mockingbird'?", 
             "What is the chemical symbol for gold?"]
prompts = [(q, f"Provide a one-word answer: {q}") for q in questions]

# 2. Experiment Execution
responses = []
for p1, p2 in prompts:
    response_p1 = openai.Completion.create(
                  engine="text-davinci-002", prompt=p1, max_tokens=3
                  )
    r1 = response_p1.choices[0].text.strip()
    response_p2 = openai.Completion.create(
                  engine="text-davinci-002", prompt=p2, max_tokens=3
                  )
    r2 = response_p2.choices[0].text.strip()
    responses.append((r1, r2))

# 3. Data Analysis
successes = 0
for r1, r2 in responses:
    len_r1 = len(r1.split())
    len_r2 = len(r2.split())
    if len_r2 <= len_r1:
        successes += 1

success_rate = successes / len(questions)

# 4. Hypothesis Testing
if success_rate > 0.5:
    print("The hypothesis is supported.")
else:
    print("The hypothesis is not supported.")

# 5. Reporting
report = pd.DataFrame(responses, columns=["R1", "R2"])
report["Success"] = report.apply(
                    lambda row: len(row["R2"].split()) \
                    <= len(row["R1"].split()), axis=1
                    )
report.to_csv("report.csv")

# 6. Review and Refinement
# This part is subjective and depends on the results of the test
\end{lstlisting}

\subsubsection{Package Install Code Generation}
\begin{lstlisting}[language=Python]
import subprocess
import sys

def install(package):
    subprocess.check_call([sys.executable, "-m", "pip", "install", package])

install('openai')
install('pandas')
\end{lstlisting}

\section{Sample Analysis}
\label{section:sample-analysis}

\subsection{Hypothesis Formulation}

\subsubsection{Problem Description}
\label{appendix:sample-analysis-problem-description}

The issue we set this time is that the LLMs outputs extraneous text other than the answer. When generating a hypothesis in response to this, as explained in Section \ref{section:results-overview}, most of the generated hypotheses were about prompt engineering. As explained in Section \ref{section:prompts-verification-plan-design}, we found that when generating a verification plan directly from this hypothesis, only vague plans were output. Therefore, we reformulated this hypothesis once and then created a verification plan, which has been converted into verification code.

Here, a problem arises as explained in Section \ref{section:control-experiment-and-verification}. That is, due to the ambiguity in the text of the hypothesis, when formulating that hypothesis, designing a verification plan, or generating a verification code based on it, its expression may deviate from the nuances of the original problem.

We will explain this issue in more detail. In the original problem statement, as shown in Listing \ref{lst:problem-text}, it is clearly stated that the extraneous text is the problem.

\begin{lstlisting}[caption={Excerpt from 
 \code{inputs/problem.txt}.},label=lst:problem-text]
...
The large language model outputs sentences that are not directly
related to the instructions.
...
\end{lstlisting}

Consequently, a hypothesis like the one in Listing \ref{lst:length-hypothesis-text} is generated. As explained in Section \ref{section:prompts-verification-plan-design}, in this example, the original phrase ``sentences that are not directly related to the instructions'' is succinctly expressed as ``concise responses''. While the former is encompassed by the latter, it is not necessarily possible to restore the former from the latter. In this sense, the nuances of the original statement are lost during the hypothesis generation stage.

\begin{lstlisting}[caption={Excerpt from \code{2023-09-0\_15-57-51/hypothesis.txt}.},label=lst:length-hypothesis-text]
...
Hypothesis: Refining the Prompting Strategy
- We could refine the way we prompt the LLM. For instance, instead
of asking "What is 1 + 1?", we could ask "Provide a one-word answer:
What is 1 + 1?". This might encourage the model to generate more
concise responses.
...
\end{lstlisting}

In hypothesis reformulation, instructions are given to model the hypothesis, namely the phrase ``concise responses''. As a result, there are cases where the modeling deviates slightly from the original nuance. For example, when given the hypothesis mentioned above, GPT-4 reformulated the hypothesis as in Listing \ref{lst:length-formulated-hypothesis-text}. That is, it represents ``concise response'' as a shorter response from the LLM.

\begin{lstlisting}[caption={Excerpt from \code{2023-09-0\_15-57-51/representation\_of\_hypothesis.txt}.},label=lst:length-formulated-hypothesis-text]
...
Mathematically, this can be represented as:
Let's denote the standard prompt as P1 and the one-word answer
prompt as P2. The responses generated by the LLM for these prompts
are denoted as R1 and R2 respectively.
The hypothesis can be tested by comparing the length (in words) of
R1 and R2. If the length of R2 is less than or equal to the length of
R1, the hypothesis is supported
...
\end{lstlisting}
In summary, what was originally described as ``sentences that are not directly related to the instructions'' was expressed as ``concise responses'' during the hypothesis generation stage. Then, during the hypothesis reformulation stage, this was interpreted as a ``short response from the LLM''. This results in an expression that differs from the original nuance.

When generating the verification plan, not only the reformulated hypothesis but also the original problem statement is input. Therefore, in principle, it is possible to retrieve the original nuance of ``concise responses'' using that information. However, in many cases, the nuance of the formulated hypothesis was directly used in the verification plan. In fact, in the verification plan when the above hypothesis model was input, the relevant part was described as in Listing \ref{lst:length-verification-plan}.

\begin{lstlisting}[caption={Excerpt from \code{2023-09-0\_15-57-51/verification\_plan.txt}.},label=lst:length-verification-plan]
...
3. Data Analysis:
3.1. For each pair of responses (R1, R2), calculate the length of
the response in words
...
\end{lstlisting}

Based on this, GPT-4 generated a verification code as shown in Listing \ref{lst:length-verification-code}.

\begin{lstlisting}[language=Python,caption={Excerpt from \code{2023-09-0\_15-57-51/verification\_code.py}.},label=lst:length-verification-code]
...
# 3. Data Analysis
successes = 0
for r1, r2 in responses:
    len_r1 = len(r1.split())
    len_r2 = len(r2.split())
    if len_r2 <= len_r1:
        successes += 1
...
\end{lstlisting}

We deemed the verification code mentioned above as valid. This is because, as explained in Section \ref{section:control-experiment-and-verification}, responses becoming shorter implies is the consequence of the LLM no longer outputting sentences unrelated to the answer. Verifying a consequence of a hypothesis doesn't directly verify the hypothesis itself. However, if the consequence is supported, it can be seen as a valid verification in the sense that it strengthens the belief that the original hypothesis seems correct.

This is structurally identical to the hypothetico-deductive method. In the hypothetico-deductive method, rather than verifying the hypothesis itself, you verify propositions deduced from the hypothesis. Since it's a verification of the deduced result, it's not a direct verification of the hypothesis. However, if the deduced result is verified, it strengthens the belief that the original hypothesis might be correct. The hypothetico-deductive method is a common practice in science. In this sense, verifying the consequence of a hypothesis doesn't seem to be an unreasonable act. This is one of the reasons we deemed the verification code mentioned above as valid.

While we made such a judgment this time, in the future, we should always be able to generate expressions of hypotheses that appropriately represent the nuances of the original problem. The primary cause of the problem this time was that the text output by hypothesis generation was too concise. Therefore, by requesting more detailed descriptions of the hypothesis either during the generation stage or after selecting from the candidates, we can expect this issue to be alleviated.

Another cause might be that during the reformulation of the hypothesis, only the hypothesis was input without the problem statement. This leads to a situation where if the generated hypothesis doesn't retain the nuances of the original problem description, it deviates from it. Thus, by also inputting the problem statement during the hypothesis formulation stage, this issue can be expected to be mitigated.

The problem this time can be seen as an issue where, once an autonomous AI deviates from the initial instructions or objectives, the deviation grows larger with each generation by the AI. This kind of problem is widely recognized when realizing autonomous AI, not limited to this instance. Therefore, advancing foundational research to fundamentally solve such issues is also one of the important challenges.

\subsubsection{Other Examples}

\subsubsubsection{\textbf{Exact Match}}

The first example examines whether the LLM's output strictly matches the answer, in accordance with the problem statement. This is the most precise representation of the hypothesis and its verification method. We have included this example in Listing \ref{lst:exact-match-verification-code}. In this example, by adding the prompt ``Provide a one-word answer:'', it investigates whether the system will strictly output only the answer. For the question ``What is 1 + 1?'', it returns 1 only if the response is ``2'', and 0 otherwise. By examining the proportion of counts it returns 1, it investigates how much the proposed method contributes to improvement. Such an example, which directly verifies the hypothesis against the problem as stated, can be considered a valid verification.

\begin{lstlisting}[language=Python,caption={Excerpt from \code{2023-09-07\_15-08-41/verfirication\_code.py}.},label=lst:exact-match-verification-code]
...
# Define the test set of prompts and expected responses
test_set = {
    "What is 1 + 1?": "2",
    "Who was the first president of the United States?": "George Washington",
}

# Define the more specific versions of the prompts
specific_prompts = {
    "What is 1 + 1?": "Provide a one-word answer: What is 1 + 1?",
    "Who was the first president of the United States?": "Provide a one-word answer: Who was the first president of the United States?",
}

# Calculate the proportion of precise responses
original_proportion = sum(original_precision) / len(original_precision)
specific_proportion = sum(specific_precision) / len(specific_precision)
...

# Evaluate the precision of the responses
original_precision = [1 if response == expected_response else 0 for response, expected_response in zip(original_responses.values(), test_set.values())]
specific_precision = [1 if response == expected_response else 0 for response, expected_response in zip(specific_responses.values(), test_set.values())]
\end{lstlisting}

\subsubsubsection{\textbf{One-Word}}
The second example is to check whether the result generated by the LLM is a single word. An example is shown in Listing \ref{lst:one-word-verification-code}. In this example, GPT-4 is not looking for an exact match with the answer. Instead, it is checking if the answer to a question that expects a one-word response is indeed one word. This evaluation does not necessarily guarantee that the generated result matches the answer, just as when checking the length. However, just as we deemed the case of length evaluation appropriate, we have determined that this is also an appropriate verification code.

\begin{lstlisting}[language=Python,caption={Excerpt from \code{2023-09-07\_16-41-46/verification\_code.py}.},label=lst:one-word-verification-code]
...
# For each response, determine whether it is a one-word answer or not
one_word_A = [1 if len(response.split()) == 1 else 0 for response in responses_A]
one_word_B = [1 if len(response.split()) == 1 else 0 for response in responses_B]
...
\end{lstlisting}

\subsubsubsection{\textbf{Numerical Response}}

The third example checks whether the outputted text is a numerical value. As explained in Section \ref{section:problem}, the problem statement uses ``What is 1 + 1?'' as an example to explain the problem. In this case, ``The answer is 2'' is a response that includes extraneous text, and the response that doesn't include any text other than the answer is ``2''. Therefore, at least for this example, checking if the answer is numerical corresponds to verifying whether the response contains only the answer. We have displayed such an example in Listing \ref{lst:numerical-verification-code}.

\begin{lstlisting}[language=Python,caption={Excerpt from \code{2023-09-07\_17-31-22/verification\_code.py}.},label=lst:numerical-verification-code]
...
# Define the Dataset
dataset = []
for _ in range(1000):
    operation = random.choice(['+', '-', '*', '/'])
    num1 = random.randint(1, 100)
    num2 = random.randint(1, 100)
    dataset.append((num1, operation, num2))

# Define the Prompts
general_prompts = [f"What is {num1} {operation} {num2}?" for num1, operation, num2 in dataset]
specific_prompts = [f"Provide the numerical answer to {num1} {operation} {num2}." for num1, operation, num2 in dataset]

# Run the Experiment
general_responses = [openai.Completion.create(engine="text-davinci-002", prompt=prompt, max_tokens=5).choices[0].text.strip() for prompt in general_prompts]
specific_responses = [openai.Completion.create(engine="text-davinci-002", prompt=prompt, max_tokens=5).choices[0].text.strip() for prompt in specific_prompts]

# Analyze the Responses
def is_numerical(response):
    return bool(re.match("^[0-9]+$", response))

general_numerical = sum(is_numerical(response) for response in general_responses)
specific_numerical = sum(is_numerical(response) for response in specific_responses)

general_percentage = (general_numerical / len(general_responses)) * 100
specific_percentage = (specific_numerical / len(specific_responses)) * 100
\end{lstlisting}

In this example, 1000 samples are mechanically and randomly generated based on the problem's example. Then, it checks whether the generated results consist only of numbers using the \code{is\_numerical} function. Of course, just checking if the answer is numerical is not strictly sufficient since it would consider answers that are only numbers but incorrect as correct. Moreover, as mentioned earlier, verifying whether the answer is numerical as a means to ensure that no extraneous text is output is limited to such examples. In this sense, this verification is not strictly sufficient. However, we judged that this verification is a somewhat valid attempt, at least within the scope of this verification, and determined that such an example is a valid verification for this time.

As also explained in Section \ref{section:control-experiment-and-verification}, the issue of the generated results being influenced by examples in the prompt is a challenge commonly observed in language models, not limited to this problem. Advancing foundational research to address such challenges is undoubtedly one of the important tasks ahead.

\subsubsubsection{\textbf{Detailed Question}}

The fourth example is about making the question more specific. As shown in Listing \ref{lst:specific-question-code}, ambiguous questions are prepared for the control group, while specific questions are prepared for the experimental group. It seems that by asking specific questions, it is expected that the answers will also become more concise.

\begin{lstlisting}[language=Python,caption={Excerpt from \code{2023-09-08\_11-33-19/verification\_code.py}.},label=lst:specific-question-code]
...
questions = [
    ("What is the weather like?", "What is the current temperature in New York?"),
    ("Tell me about dogs.", "What is the average lifespan of a Labrador Retriever?"),
    ("What's happening in the world?", "What are the current top news headlines?"),
]
...
\end{lstlisting}

We have determined that the verification code in this example is not a valid verification code. This is because, in addition to the expression of the hypothesis being different from the original intent of the problem, it neither results from the hypothesis nor holds true for the specific example given in the problem. However, it is not inherently wrong to make the question more specific. In such cases, if any of the above requirements are met, we have determined it to be a valid verification.

Regarding this example, as shown in Listing \ref{lst:specific-reformulated-hypothesis}, it was outputting valid content up to the stage of hypothesis reformulation

\begin{lstlisting}[caption={Excerpt from \code{2023-09-08\_11-33-19/representation\_of\_hypothesis.txt}.},label=lst:specific-reformulated-hypothesis]
...
The specificity of a question prompt directly influences the precision of the response. For instance, modifying a general question like "What is 1 + 1?" to a more specific one such as "Provide the numerical answer to 1 + 1" will yield a more precise numerical answer.
...
\end{lstlisting}

However, as shown in Listing \ref{lst:specific-verification-plan}, the expression of the hypothesis became ambiguous again at the verification plan stage. The verification plan of this example itself is generally valid. However, due to the ambiguity remaining in the part shown below, it is speculated that a verification code that deviates from the nuance of the original problem was generated

\begin{lstlisting}[caption={Excerpt from \code{2023-09-08\_11-33-19/verification\_plan.txt}.},label=lst:specific-verification-plan]
...
1. Define the Experiment:
   1.1. Define a set of questions that will be used in the experiment. These questions should be diverse and cover a range of topics to ensure the results are not biased towards a specific type of question.
   1.2. For each question, create two versions: a general version and a specific version. The specific version should be designed to elicit a more precise response according to the hypothesis.
...
\end{lstlisting}

\subsection{Ellipsis, Placeholder, and Comments}
In some cases, there were instances where GPT-4s did not create data or functions themselves. Specifically, there were times when areas that should have had values were instead terminated with just ellipsis, placeholders, or comments. We have displayed the parts corresponding to each in Listing \ref{lst:ellipsis}, \ref{lst:placeholder}, and \ref{lst:comments}.

\begin{lstlisting}[language=Python,caption={Excerpt from \code{2023-09-08\_10-26-23/verification\_code.py}.},label=lst:ellipsis]
...
general_prompts = [...]  # Replace with your general prompts
specific_prompts = [...]  # Replace with your specific prompts
...
\end{lstlisting}

\begin{lstlisting}[language=Python,caption={Excerpt from \code{2023-09-07\_17-31-22/verification\_code.py}.},label=lst:placeholder]
...
def ask_llm(question):
    return "Placeholder response"

general_responses = [ask_llm(q) for q in general_questions]
specific_responses = [ask_llm(q) for q in specific_questions]
...
\end{lstlisting}

\begin{lstlisting}[language=Python,caption={Excerpt from \code{2023-09-07\_16-21-44/verification\_code.py}.},label=lst:comments]
...
math_questions = [
    # Add your list of mathematical questions here
    # Each question should be a tuple with two elements:
    # The first element is the non-specific version of the question
    # The second element is the specific version of the question
    # For example: ("What is 1 + 1?", "Provide the numerical answer to 1 + 1")
]
...
\end{lstlisting}
In the example of Listing \ref{lst:ellipsis}, instead of defining sample questions, only ellipsis is outputted. In the example of Listing \ref{lst:placeholder}, the part to obtain output from the language model is only outputted as a placeholder, without actually using libraries like \code{openai}. In Listing \ref{lst:comments}, even though the necessary actions are understood, they are only outputted as comments.

This behavior is likely a result of the language model being adjusted to produce such outputs during its training process, or due to the presence of many such examples in the training data. Refraining from outputting beyond its capabilities and seeking human guidance is desired behavior for a human assistant. Therefore, this probably won't be an issue when using the language model as an auxiliary tool for code generation. However, it becomes a problem when expecting the AI to autonomously conduct research without human intervention. If this behavior is a result of the training process, updating the training phase will be necessary to create an autonomous AI.

We determined that such results are inappropriate as verification code content for this time. This is because our objective was to investigate whether the AI itself can conduct research autonomously, and relying on human intervention at any stage means it cannot be called autonomous hypothesis verification by the AI. Instead, even if the output contained some errors, had limited data, or seemed toy-like, as long as the AI produced results based on its own reasoning and they weren't too absurd for human understanding, we overlooked such mistakes. This is because they are results of the AI's attempt to conduct research autonomously.

\subsection{API Key}
A frequently observed undesirable behavior related to placeholders was the instance where the API Key was substituted with \code{your\_api\_key}. I have shown this example in Listing \ref{lst:apikey}. Even if the API Key is defined locally, having this in the code will render it non-functional.

\begin{lstlisting}[language=Python,caption={Excerpt from \code{2023-09-07\_16-21-44/verification\_code\_updated.py}.},label=lst:apikey]
...
# Initialize OpenAI
openai.api_key = 'your-api-key'
...
\end{lstlisting}
Due to the frequent occurrence of this issue, as explained in Section \ref{section:hypothesis-verification-module}, after generating the verification code once, we instructed GPT-4 to exclude the API key. We made sure the instruction was not limited to just the \code{openai} API Key. The actual prompt is shown in Figure \ref{fig:prompt-instruction-following}. As indicated by the phrase ``DO NOT include api-key in the code, as it has already been specified'', we ensured that no API keys, not just the OpenAI API key, were specified.

\subsection{Statistical Hypothesis Test}

As mentioned in Section \ref{section:control-experiment-and-verification}, statistical hypothesis testing was attempted in many cases. Specifically, over the half of the total cases attempted statistical hypothesis testing.

I've provided an example in Listing \ref{lst:hypothesis_test}. In this example, \code{general\_lengths} represents the control group, and \code{specific\_lengths} represents the experimental group. Using these two, an independent t-test is conducted, and if the p-value is less than 0.05, it concludes that the hypothesis was valid. Since statistical hypothesis testing is widely used in empirical science to judge verification results, the fact that the AI can autonomously use statistical hypothesis testing is a promising outcome.

\begin{lstlisting}[language=Python,caption={Excerpt from \code{2023-09-07\_18-28-53/verification\_code.py}.},label=lst:hypothesis_test]
...
from scipy import stats
...
# Statistical testing
t_stat, p_val = stats.ttest_ind(general_lengths, specific_lengths)

# Result interpretation
if p_val < 0.05:
    print("The specificity of a question prompt directly influences the conciseness of the response.")
else:
    print("The specificity of a question prompt does not directly influence the conciseness of the response.")
\end{lstlisting}

On the other hand, there are several challenges. For instance, to conduct a t-test, one must check for independence, homogeneity of variance, and normality. However, in this example, these checks are not performed (the \code{scipy.stats.ttest\_ind} function assumes equal variance by default). Also, as mentioned in Section \ref{section:results-and-discussion-generate-results}, given the small sample size, it's unrealistic to expect meaningful results from statistical hypothesis testing. Furthermore, this test conducts a two-tailed test (the \code{scipy.stats.ttest\_ind} function conducts a two-tailed test by default). However, since we want to verify whether the proposed method shortened the output length of the language model, a one-tailed test would be more appropriate. From these observations, it seems that the AI does not fully understand and appropriately utilize hypothesis testing.

In this study, even if such inappropriate hypothesis tests were included, if the overall verification process seemed generally valid, we deemed the verification code as appropriate. This judgment is based on the fact that, given no specific instructions on the method of verification, arriving at the idea of ``using hypothesis testing'' and attempting its use is sufficiently valid as a verification attempt. Moreover, while prerequisites like independence, homogeneity of variance, and normality should undoubtedly be checked, humans unfortunately often use hypothesis testing without verifying these conditions. It would be harsh to demand from AI what humans often fail to do, so we decided to tolerate such flaws for this study.

The primary focus of this study was to investigate whether the AI could attempt verification on its own. To ensure that the AI can conduct strictly appropriate verifications, foundational research aiming to overcome the challenges identified in this study seems necessary.

\subsection{Errors Related to OpenAI API}
As mentioned in Section \ref{section:results-and-discussion-generate-results}, errors related to the use of OpenAI's API were frequently observed. For instance, there were errors like in Listing \ref{lst:openai_api_1} and \ref{lst:openai_api_2}, where the code attempts to call something that doesn't exist in OpenAI. There were also cases like in Listing \ref{lst:openai_api_3}, where the \code{engine} was either not specified or an incorrect one was specified.

\begin{lstlisting}[language=Python,caption={Excerpt from \code{ 2023-09-07\_16-16-40/verification\_code.py}.},label=lst:openai_api_1]
...
# Initialize the LLM
llm = openai.LanguageModel()
...
\end{lstlisting}

\begin{lstlisting}[language=Python,caption={Excerpt from \code{2023-09-08\_11-36-40/verification\_code.py}.},label=lst:openai_api_2]
...
from openai import GPT3
...
\end{lstlisting}

\begin{lstlisting}[language=Python,caption={Excerpt from \code{2023-09-08\_10-57-56//verification\_code.py}.},label=lst:openai_api_3]
...
# Conduct the experiment
def conduct_experiment(prompts):
    responses = []
    for prompt in prompts:
        response = openai.Completion.create(engine="davinci-codex", prompt=prompt, max_tokens=5)
        responses.append(response.choices[0].text.strip())
    return responses
...
\end{lstlisting}

As mentioned in Section \ref{section:results-and-discussion-generate-results}, many of these issues could be quickly resolved with code updates based on the errors. Since \code{GPT-4} has only been trained on data up to 2021, such errors occur. However, currently, more people are using \code{openai}. Therefore, it's expected that these errors will decrease sooner, so it doesn't seem to be a significant concern. Therefore, we have determined that if there are errors related to OpenAI's API, but the rest of the content is valid, then the verification code is considered appropriate.

\end{document}